\documentclass[11pt]{article}

\usepackage[preprint]{acl}

\usepackage{times}
\usepackage{latexsym}

\usepackage[T1]{fontenc}

\usepackage[utf8]{inputenc}

\usepackage{microtype}

\usepackage{inconsolata}

\usepackage{graphicx}

\usepackage{url}
\usepackage{booktabs}
\usepackage{multirow}
\usepackage{xcolor}
\usepackage{colortbl}
\usepackage{wrapfig}
\usepackage{pifont}
\usepackage{amsmath}
\usepackage{makecell}
\usepackage{wasysym}
\usepackage{amssymb}
\usepackage{alphalph}
\usepackage{subcaption}

\title{Chart-R1: Chain-of-Thought Supervision and Reinforcement\\ for Advanced Chart Reasoner}

\author{Lei Chen \quad Xuanle Zhao \quad Zhixiong Zeng$^{\dag}$ \quad Jing Huang \quad Yufeng Zhong \quad Lin Ma$^{\ast}$ \\
Meituan \\
\texttt{leichen1997@outlook.com \quad zengzhixiong@meituan.com \quad forest.linma@gmail.com}
}

\begin{document}
\maketitle

{\let\thefootnote\relax\footnotetext{$^{\dag}$ Project leader. $^{\ast}$ Corresponding author.}}

\begin{abstract}

Chart reasoning presents unique challenges due to its inherent complexity—requiring precise numerical comprehension, multi-level visual understanding, and logical inference across interconnected data elements. Existing vision-language models often struggle with such reasoning tasks, particularly when handling multi-subchart scenarios and numerical sensitivity. To address these challenges, we introduce \textbf{Chart-R1}, a chart-domain vision-language model that leverages reinforcement fine-tuning for advanced chart reasoning. We first propose a programmatic data synthesis approach to generate high-quality step-by-step reasoning data with verifiable answer formats, covering diverse chart types and complexity levels. Our two-stage training strategy includes: (1) \textbf{Chart-COT}, which decomposes complex reasoning into interpretable subtasks through chain-of-thought supervision, and (2) \textbf{Chart-RFT}, which employs group relative policy optimization with numerically sensitive rewards tailored for chart-specific reasoning. Experiments on open-source benchmarks and our proposed \textbf{ChartRQA} dataset demonstrate that Chart-R1 significantly outperforms existing chart-domain methods and rivals large-scale open/closed-source models.
\end{abstract}

\section{Introduction}
\label{sec:intro}

Recently, inspired by the success of models such as OpenAI's o1/o3 \cite{openai2025o3o4mini} and DeepSeek-R1 \cite{guo2025deepseek}, leveraging Reinforcement Learning (RL) for fine-tuning has garnered significant attention within the research community. Although these methods have shown promise in textual domains like mathematical reasoning, code generation, and multidisciplinary knowledge,  transferring these advanced reasoning capabilities to the vision domain presents an open challenge. While recent approaches like Vision-R1 \cite{huang2025vision} and VLM-R1 \cite{shen2025vlm} have successfully leveraged RL to enhance visual perception and grounding,  they have primarily focused on simple questions, neglecting tasks that demand deep reasoning capabilities.

Charts, as information-intensive images, are a crucial research area in image understanding and reasoning \cite{wang2024charxiv}. Chart reasoning requires decomposing complex questions into multiple interpretable steps involving precise numerical comprehension, visual parsing, and logical inference. As illustrated in Figure~\ref{fig:res_comp_1}, questions with fewer reasoning steps are relatively straightforward, but the difficulty increases substantially when problems require multi-step decomposition.
Prior works attempt to improve chart reasoning through supervised fine-tuning (SFT) on datasets augmented with Chain-of-Thought (CoT) or Program-of-Thought (PoT) methods \cite{wei2022chain,chen2022program}. However, SFT causes models to overfit specific reasoning patterns, hindering their generalization abilities. Recent methods \cite{jia2025chartreasoner, ni2025point} leverage RL to enhance VLM reasoning capabilities, yet their focus remains primarily on perception and understanding rather than the multi-step reasoning required for complex chart analysis.

 
\begin{figure*}[t]
    \centering
    \includegraphics[width=1.0\linewidth]{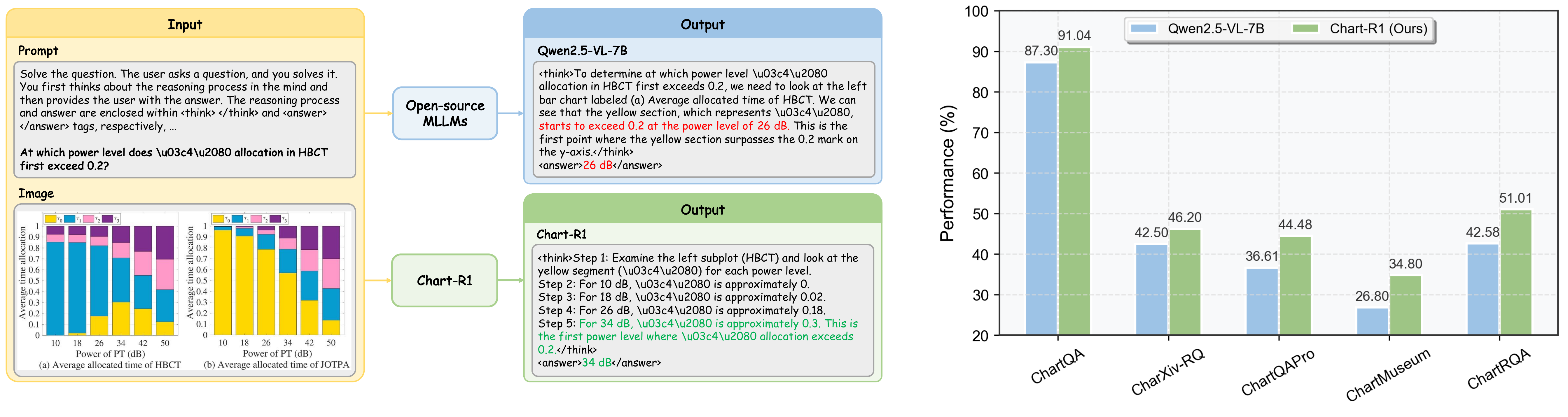}
    \vspace{-25pt}
    \caption{Comparison of Qwen2.5-VL-7B and Chart-R1 on chart understanding and reasoning benchmarks. In the complex chart reasoning task, Qwen2.5-VL-7B generates a wrong thinking process, whereas Chart-R1 thinks and answers correctly.}
    \label{fig:intro}
    \vspace{-10pt}
\end{figure*}

In this work, we propose Chart-R1, a chart-domain VLM that leverages RL to enhance advanced multi-step reasoning capability, which achieves superior performance as shown in Figure~\ref{fig:intro}. To support Chart-R1, we introduce two key contributions. 
First, we propose a programmatic synthesis strategy to generate high-quality step-by-step reasoning data with verifiable answer formats, covering diverse chart types and complexity levels. Specifically, we utilize LLMs to generate the chart plotting code from real-world tables curated from arXiv papers, ensuring data fidelity in the resulting charts. The generated code then serves as a foundation for formulating complex questions, multi-step CoT reasoning processes, and verifiable answers, resulting in ChartRQA—a comprehensive dataset featuring 258k multi-step reasoning samples covering both single- and multi-chart scenarios.
Second, we develop an effective two-stage training strategy: (1) Chart-COT with step-by-step chain-of-thought supervision, and (2) Chart-RFT with numerically sensitive reinforcement fine-tuning. During the Chart-COT stage, the model learns to decompose complex tasks into fine-grained, interpretable subtasks through supervised learning on step-by-step reasoning data. As shown in Figure~\ref{fig:intro}, this explicit decomposition capability is crucial for handling complex scenarios that require multi-step reasoning.
In the Chart-RFT stage, we employ group relative policy optimization (GRPO) with a composite reward signal of soft matching and edit distance to enhance accuracy for both numerical and string-based answers. We utilize distinct datasets for these two stages, as our findings show that training on the same data impairs the model's exploration ability during the RL process.
Furthermore, we introduce ChartRQA, a human-verified benchmark designed to probe the limits of complex chart reasoning. In contrast to prior works \cite{wang2024charxiv}, its questions demand higher complexity and multi-step thought processes. The substantial performance drop of existing VLMs on ChartRQA reveals a critical gap in their reasoning capabilities. In summary, our contributions are as follows:
\begin{itemize}
    \item To enhance chart reasoning in VLMs, we propose a novel two-stage training strategy consisting of Chart-COT and Chart-RFT. The resulting model, Chart-R1, sets a new state-of-the-art across various chart understanding and reasoning benchmarks.
    \item We introduce a programmatic data synthesis strategy that leverages code generation from real-world tables to produce step-by-step reasoning data with verifiable answers, ensuring both data fidelity and reasoning quality.
    \item We present ChartRQA, a comprehensive dataset with a human-verified benchmark and 258k training samples. Existing VLMs show significant performance gaps on this benchmark, revealing critical reasoning limitations.
\end{itemize}

\section{Related Works}
\label{sec:related_work}

\subsection{Chart VLMs}
Chart understanding and reasoning are crucial areas of research community that encompass both low-level and high-level tasks \cite{singh2019towards, methani2020plotqa}. Recently, many chart-domain models have been proposed to enhance the chart understanding capacity of VLMs \cite{han2023chartllama, liu2023mmc}. However, prior works concentrate on descriptive tasks \cite{masry2024chartinstruct, masry2024chartgemma}, such as extracting explicit content from charts \cite{masry2022chartqa}. More recent works focus on leveraging reasoning capabilities to interpret complex and implicit information. For example, TinyChart \cite{zhang2024tinychart} utilizes a template-based method to generate Program-of-Thought (PoT) \cite{chen2022program} reasoning data. ChartCoder \cite{zhao2025chartcoder} proposes Snippet-of-Thought for chart-to-code generation. ChartReasoner \cite{jia2025chartreasoner} utilizes a chart-to-code model to convert images into code and generate reasoning processes. However, the generated reasoning data has limitations due to chart-to-code accuracy \cite{shi2024chartmimic, xu2024chartmoe}.

\subsection{Long Reasoning VLMs}
Recently, with the success of DeepSeek-R1 \cite{guo2025deepseek}, many works attempt to enhance LLM reasoning ability via rule-based reward and RL \cite{shao2024deepseekmath}. In the vision-language domain, recent works follow the DeepSeek-R1 method to enhance the long-chain reasoning capacity of VLMs \cite{shen2025vlm, wang2025multimodal,qiu2025metis}. For example, Vision-R1 \cite{huang2025vision} and R1-OneVision \cite{yang2025r1} apply Group Relative Policy Optimization (GRPO) with multimodal reasoning data to enable long reasoning. MMEureka \cite{meng2025mm} and R1-Zero \cite{liu2025understanding} further advance visual long-term reasoning with improved RL training strategies. Point-RFT \cite{ni2025point} uses grounded CoT for visual understanding, but it only utilize ChartQA for RL which limits the model's reasoning capacity.

\begin{figure*}[t]
    \centering
    \includegraphics[width=1.0\linewidth]{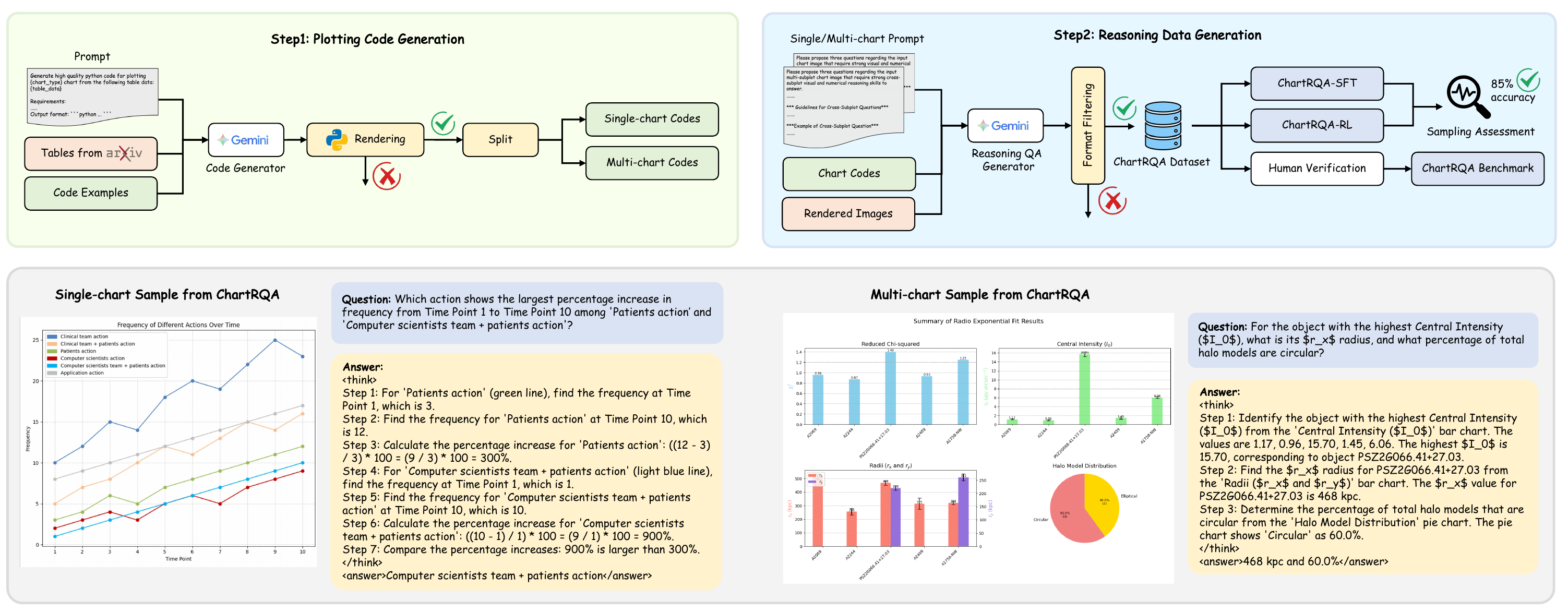}
    \vspace{-15pt}
    \caption{ChartRQA dataset pipeline via programmatic data synthesis. The pipeline consists of plotting code generation and reasoning data generation, producing ChartRQA-SFT, ChartRQA-RL, and the ChartRQA benchmark. Examples show single-chart and multi-chart scenarios requiring multi-step reasoning.}
    \label{fig:data_pipe}
    \vspace{-5pt}
\end{figure*}

\subsection{Chart Understanding and Reasoning}
A variety of training datasets and evaluation benchmarks have been developed to improve VLM performance on chart-related tasks \cite{xia2024chartx, shi2024chartmimic, he2024distill,zhao2025chartedit, wu2025chartcards}. Previous works generally focus on description tasks. For example, ChartQA \cite{masry2022chartqa}, PlotQA \cite{methani2020plotqa} and Chart-to-text \cite{kantharaj2022chart} mainly train and evaluate models' capacity on extracting information from charts, where the challenge is predominantly driven by chart complexity. Recent works such as Charxiv \cite{wang2024charxiv} and CharMuseum \cite{tang2025chartmuseum} introduce more challenging reasoning tasks that require models to think before answering. Unlike descriptive tasks, reasoning tasks present a dual challenge from both the perceptual complexity of charts and the reasoning depth required by questions.

\section{Method}
\label{sec:method}

To enhance the reasoning capabilities of models on chart reasoning tasks, we introduce our proposed data synthesis and two-stage training strategy. We first programmatically generate a large-scale training dataset with the CoT reasoning process and subsequently employ the SFT on CoT data as a cold start phase to bootstrap the subsequent RL strategy for training.

\begin{table*}[t]
  \centering
  \small
  \begin{tabular}{l|ccccc}
    \toprule
    \textbf{Dataset} & \textbf{Types} & \textbf{Unique Charts} & \textbf{Multi-chart} & \textbf{Thinking Process} \\  
    \midrule
    ChartQA~\cite{masry2022chartqa} & 3 & 21.9k & \ding{55} & \ding{55} \\
    MMC~\cite{liu2023mmc} & 7 & 600k & \ding{52} & \ding{55} \\
    ChartLlama~\cite{han2023chartllama} & 10 & 11k & \ding{55} & \ding{55} \\
    NovaChart~\cite{hu2024novachart} & 18 & 47k & \ding{55} & \ding{52}\rotatebox[origin=c]{-9.2}{\kern-0.7em\ding{55}} \\
    ECD~\cite{yang2025effective} & 29 & 10.5k & \ding{52} & \ding{55}\\
    ChartRQA (Ours) & 24 & 93.3k & \ding{52} & \ding{52} \\
    \bottomrule
  \end{tabular}
  \caption{Comparison of our proposed ChartRQA training set with other chart datasets. ChartRQA features the integration of single/multi-charts, thinking processes, and verifiable answer formats.}
  \label{dataset_comp}
\end{table*}

\begin{table}[t]
  \centering
  \small
  \begin{tabular}{l|cc|cc}
    \toprule
    \multirow{2}{*}{\textbf{Token Avg.}} & \multicolumn{2}{c|}{\textbf{Train}} & \multicolumn{2}{c}{\textbf{Test}} \\
    & Single & Multi & Single & Multi \\
    \midrule
    Question & 30.03 & 39.84 & 29.83 & 39.49 \\
    Thinking Process & 196.50 & 237.38 & 196.32 & 240.94 \\
    Answer & 5.98 & 8.87 & 5.96 & 8.97 \\
    \bottomrule
  \end{tabular}
  \caption{The average question, thinking process, and answer lengths in the ChartRQA train and test sets. We count the single- and multi-chart problems of each set separately.}
  \label{avg_token_length}
  \vspace{-15pt}
\end{table}

\subsection{Programatic Data Synthesis}
\label{sec:dataset}
While several CoT datasets for chart reasoning have been proposed, they are largely derivatives of the ChartQA dataset, constructed by augmenting its existing question-answer pairs with generated reasoning processes \cite{zhang2024tinychart, jia2025chartreasoner}.
However, this method is like distilling reasoning from top VLMs, so it naturally inherits their limitations and errors on complex tasks.
The reliance on final answer correctness as the only supervisory signal makes generating high-quality CoT reasoning data a significant challenge. This issue is amplified in complex chart reasoning, where the struggles of even top models inherently lead to low-quality, undiverse data.
Although the recent ChartReasoner method \cite{jia2025chartreasoner} generates reasoning data by first parsing charts into code, the diversity and quality of generated data are fundamentally limited by the performance of the chart-to-code parser.
In contrast, our programmatic data generation strategy reverses this paradigm by utilizing code as a pivotal starting source. First, we prompt a powerful LLM to generate plotting code. This code then serves as a perfect, high-fidelity foundation from which a VLM subsequently synthesizes question-answer pairs and their complex step-by-step reasoning path. An overview of our data synthesis pipeline is shown in Figure~\ref{fig:data_pipe}.

\textbf{Plotting Code Generation} We instruct LLMs to generate Matplotlib plotting code to render high-quality and diverse chart images. 
However, our analysis reveals that directly generating synthetic data values in plotting code often yields monotonous trends that lack complexity and diversity. To address this, we first curate tables from real-world arXiv papers, which serve as veritable data sources. Secondly, to enhance the diversity of the generated code, we manually write seed code examples for different chart types. To ensure the diversity of generated code, we randomly combine the curated table and seed code as in-context learning sources for LLMs to generate plotting code.
To generate complex, multi-chart scenarios, we both include numerous multi-chart examples in our seed code and explicitly prompt the LLM during generation to use functions like plt.subplots() to create composite figures. Our work significantly expands the range of chart types available for chart reasoning, representing the most diverse dataset. We execute all generated code samples and discard any that fail to run successfully.

\textbf{Reasoning Data Generation} With the executable plotting code as a foundation, we prompt LLMs to synthesize a complete reasoning instance, comprising a question, its answer, and a step-by-step reasoning path. To enhance diversity, we categorize the plotting code into single- and multi-chart types and apply distinct generation instructions for each. For multi-chart problems, we instruct the LLM to generate questions that necessitate cross-referencing information between sub-charts. The generated data show that this strategy significantly enhances multi-chart task complexity. Our results show that leveraging code allows LLMs to produce more complex questions and detailed reasoning compared to methods that use chart images alone. We posit that a code-based approach is superior for generating complex chart reasoning as the underlying code provides a lossless textual representation while enabling the scalable synthesis of data independent of existing corpora. 

\textbf{Dataset Construction}
Using the aforementioned methods, we construct the ChartRQA dataset, which includes a large-scale training dataset of 258k instances with reasoning paths as well as a human-verified benchmark. The training dataset is separated into two subsets for our two-stage training strategy, ChartRQA-SFT and ChartRQA-RL, consisting of 228k and 30k samples, respectively. 
Detailed comparisons about ChartRQA with other chart-domain training sets are denoted in Table~\ref{dataset_comp}. 
The benchmark is constructed via a human validation where experts review each sample for question difficulty and answer correctness, subsequently constructing 1,702 high-quality samples (933 single-chart and 769 multi-chart tasks) for evaluation. As detailed in Table~\ref{avg_token_length}, we also calculated the average token counts for the questions, reasoning paths, and final answers, broken down by single- and multi-chart problems. The analysis reveals that the components associated with multi-chart problems are significantly longer than those for single-chart problems. 

\textbf{Quality Evaluation}
To assess the quality of our generated data, we randomly sample 1k instances and recruit human experts for evaluation. The results indicate that over 85\% of the instances are free from errors. Notably, we deliberately omit any data cleaning process. The fact that our model, Chart-R1, achieves strong performance despite being trained on this raw, uncurated dataset validates the robustness of our proposed code-based generation strategy.

\subsection{Chart-COT}
To enhance the chart reasoning capacity, we propose a two-stage training strategy. Using Qwen2.5-VL-7B-Instruct as the baseline model, we first fine-tune it via SFT on the step-by-step reasoning data of our proposed ChartRQA-SFT. 
This initial stage equips the model with the fundamental capability to decompose complex tasks into fine-grained subtasks.
Our ablation studies demonstrate that this preliminary SFT stage on CoT data is critical, as it yields significantly better performance than applying RL from scratch.
The model is trained using a standard autoregressive language modeling objective with negative log-likelihood loss.

\subsection{Chart-RFT}
After the Chart-COT stage, while the fine-tuned model demonstrates an enhanced ability to decompose complex questions, its performance on out-of-domain (OOD) tasks notably degrades. We hypothesize this is due to a distributional mismatch between ChartRQA-SFT with some simple chart understanding tasks, which harms its generalization ability. To address this degradation, we subsequently apply reinforcement fine-tuning (RFT) to generalize its reasoning capacity.

\textbf{Group Relative Policy Optimization} 
We adapt the Group Relative Policy Optimization (GRPO) algorithm \cite{shao2024deepseekmath,guo2025deepseek}, which significantly conserves training resources by replacing the critic model with a baseline estimated from group scores.
For each input $(x, y)$, the policy $\pi_\theta$ samples a group of $G$ candidate responses $\{o_i\}_{i=1}^G$.
\begin{align}
\mathcal{J}_{GRPO}(\theta) = & \mathbb{E}_{\substack{\{o_i\}_{i=1}^G \sim \pi_{\theta_{\text{old}}} (O \mid x)}} \notag \\
&\biggl[ \frac{1}{G} \sum_{i=1}^G \min \biggl(\frac{\pi_\theta(o_i \mid x)}{\pi_{\theta_{\text{old}}}(o_i \mid x)} A_i,\\
& \operatorname{clip}\left(\frac{\pi_\theta(o_i \mid x)}{\pi_{\theta_{\text{old}}}(o_i \mid x)}, 1-\varepsilon, 1+\varepsilon\right) A_i\biggr) \notag \biggr]
\end{align}
where $\varepsilon$ is the hyperparameter, $\pi_{\theta}$ and $\pi_{\theta_{\text{old}}}$ are the optimized model and the policy model respectively.
The group-normalized advantage for the $i$-th response is:
\begin{equation}
A_i=\frac{r_i-\operatorname{mean}\left(\left\{r_1, r_2, \cdots, r_G\right\}\right)}{\operatorname{std}\left(\left\{r_1, r_2, \cdots, r_G\right\}\right)}
\end{equation}

\textbf{Reward Design}
For effective RFT, we follow the DeepSeek-R1 \cite{shao2024deepseekmath} and adopt a rule-based reward function that consists of accuracy and format rewards to assess answer correctness and structural integrity, respectively.

\begin{itemize}
    \item \textbf{Accuracy Reward.} We employ distinct, type-specific reward functions to measure the correctness of model outputs. For numerical answers, we adopt the soft matching technique from Point-RFT \cite{ni2025point} with a relative error tolerance of $\pm$5\%. For string-based answers, we utilize the edit distance as the reward signal.
    \item \textbf{Format Reward.} The format reward is determined by a grammar-level regex parser that validates the structural integrity of outputs. It confirms two conditions: (1) the reasoning process is properly enclosed in $<$think$>$ tags, and (2) the final answer is extractable from the designated $<$answer$>$ tags.
\end{itemize}

\textbf{Data Proportion}
For the Chart-COT and Chart-RFT stages, we utilize distinct subsets of ChartRQA. This setting is critical, as our experiments reveal that using the same CoT data for both phases causes the model to overfit to replicate the reasoning paths from the SFT data, which in turn degrades the diversity and exploration capability of the policy model during the RL phase. 
We find that the stability and convergence of the Chart-RFT phase critically depend on the pattern consistency of the data from the preceding Chart-COT stage. Employing SFT data with inconsistent patterns significantly hinders RFT convergence, highlighting the necessity of a distributionally aligned dataset in the Chart-COT stage to ensure effective downstream RFT.

\begin{table*}[t]
  \centering
  \small
  \begin{tabular}{l|ccccc}
    \toprule
    \textbf{Model Name} & \textbf{ChartQA} & \textbf{CharXiv-RQ} & \textbf{ChartQAPro} & \textbf{ChartMuseum} & \makecell{\textbf{ChartRQA}\\(single / multi)} \\
    \midrule
    \multicolumn{6}{c}{\textit{Proprietary}} \\
    \midrule GPT-4o & 85.7 & 47.1 & 37.67 & 42.2 & 44.37 / 46.55 \\
    Gemini-1.5-Flash & 79.0 & 33.9 & 42.96 & 31.1 & - \\
    Gemini-1.5-Pro & 87.2 & 43.3 & - & 41.3 & - \\
    Gemini-2.5-Flash & - & - & - & - & 59.12 / 59.17 \\
    Claude-3.5-Sonnet & 90.8 & 60.2 & 43.58 & 54.4 & 52.79 / 56.05 \\
    GPT-4.1 & 86.8 & 56.7 & - & 48.4 & 57.88 / 59.30 \\
    Claude-3.7-Sonnet & 86.1 & 64.2 & - & 60.3 & 55.04 / 57.87 \\
    \midrule
    \multicolumn{6}{c}{\textit{General-domain Open-source}} \\
    \midrule Phi-3.5-Vision & 81.8 & 32.7 & 24.73 & - & 31.08 / 24.32 \\
    DeepSeek-VL2 & 86.0 & - & 16.28 & - & 23.15 / 20.29 \\
    InternVL3-8B  & 86.6 & 37.6 & - & 28.2 & 37.51 / 31.73 \\
    InternVL3-38B  & 89.2 & 46.4 & - & 32.1 & 46.09 / 38.36 \\
    Qwen2.5-VL-7B & 87.3 & 42.5 & 36.61 & 26.8 & 44.59 / 40.57 \\
    R1-VL & 83.9 & 32.0 & 35.11 & 21.0 & 27.33 / 21.72 \\
    VL-Rethinker & 83.5 & 42.8 & \textbf{48.26} & 31.5 & 42.87 / 42.52 \\
    MMR1 & 83.7 & 40.5 & 45.72 & 30.9 & 47.27 / 46.55 \\
    \midrule
    \multicolumn{6}{c}{\textit{Chart-domain}} \\
    \midrule ChartLlama & 69.66 & 14.2 & - & - & - \\
    TinyChart & 83.60 & 8.3 & 13.25 & 12.5 & 6.75 / 6.11 \\
    ChartGemma & 80.16 & 12.5 & 6.84 & 12.2 & 7.18 / 9.23 \\
    ChartReasoner & 86.93 & - & 39.97 & - & - \\
    BigCharts-R1 & 89.84 & 41.3 & - & - & - \\
    Bespoke-MiniChart & 89.50 & 45.4 & 45.36 & 34.0 & 42.77 / 42.13 \\
    Chart-R1 (Ours) & \textbf{91.04} & \textbf{46.2} & 44.48 & \textbf{34.8} & \textbf{52.09} / \textbf{49.93} \\
    \bottomrule
  \end{tabular}
  \caption{The main results on existing chart understanding and reasoning benchmarks. Our proposed Chart-R1 achieves superior performance among small-scale VLMs ($<$20B) on the evaluation benchmarks. \textbf{Bold} denotes the best performances of open-source VLMs.}
  \label{sota}
\vspace{-5pt}
\end{table*}

\section{Experiments}
\label{sec:experiment}

\subsection{Experiment Settings}
We conduct experiments and ablation studies to evaluate the results obtained from various training settings. See the appendix for training details.

\textbf{Benchmarks}
To comprehensively evaluate the understanding and reasoning capacity of Chart-R1, we choose ChartQA \cite{masry2022chartqa}, CharXiv-RQ (Reasoning Questions) \cite{wang2024charxiv}, ChartQAPro \cite{masry2025chartqapro}, ChartMuseum \cite{tang2025chartmuseum} and our proposed ChartRQA (single/multi) as the evaluation benchmarks.

\textbf{Baselines} 
We compare our proposed Chart-R1 with existing models in three setups: (1) Proprietary models include GPT-4o, GPT-4.1 \cite{openai2025gpt41}, Gemini-1.5-(Flash, Pro), Gemini-2.5-Flash \cite{kavukcuoglu2025gemini}, Claude-3.5-Sonnet and Claude-3.7-Sonnet \cite{anthropic2025claude37}. 
(2) General-domain open-source VLMs, including 
Phi 3.5-Vision \cite{abdin2024phi3technicalreport}, DeepSeek-VL2 \cite{wu2024deepseekvl2}, 
InternVL3(8B, 38B) \cite{zhu2025internvl3} and Qwen2.5-VL(7B) \cite{bai2025qwen25}; and VLM reasoning models, including R1-VL \cite{zhang2025r1}, VL-Rethinker \cite{wang2025vl} and MMR1 \cite{leng2025mmr1}. 
(3) Chart-domain VLMs including ChartLlama \cite{han2023chartllama}, TinyChart \cite{zhang2024tinychart},  ChartGemma \cite{masry2024chartgemma}, ChartResoner \cite{jia2025chartreasoner}, BigCharts-R1 \cite{masry2025bigcharts} and Bespoke-MiniChart-7B \cite{bespoke_minichart_7b}.

\begin{figure*}[t]
    \centering
    \includegraphics[width=0.9\linewidth]{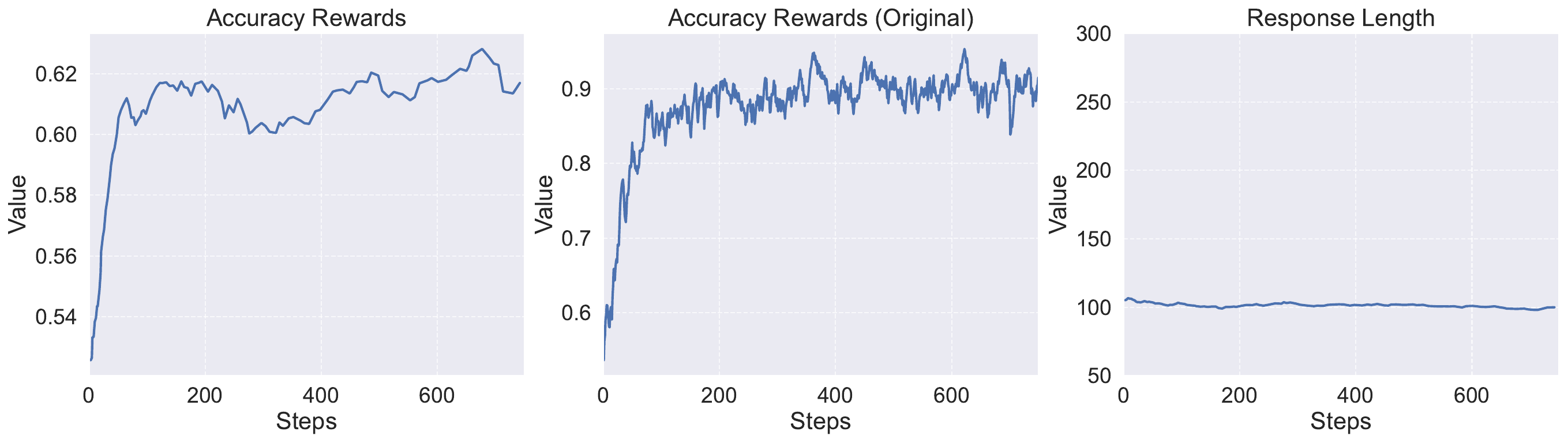}
    \vspace{-5pt}
    \caption{The training curve during the RL stage that utilizes the ChartQA dataset solely.}
    \label{fig:rl_chartqa}
    \vspace{-5pt}
\end{figure*}

\begin{figure*}[t]
    \centering
    \includegraphics[width=0.9\linewidth]{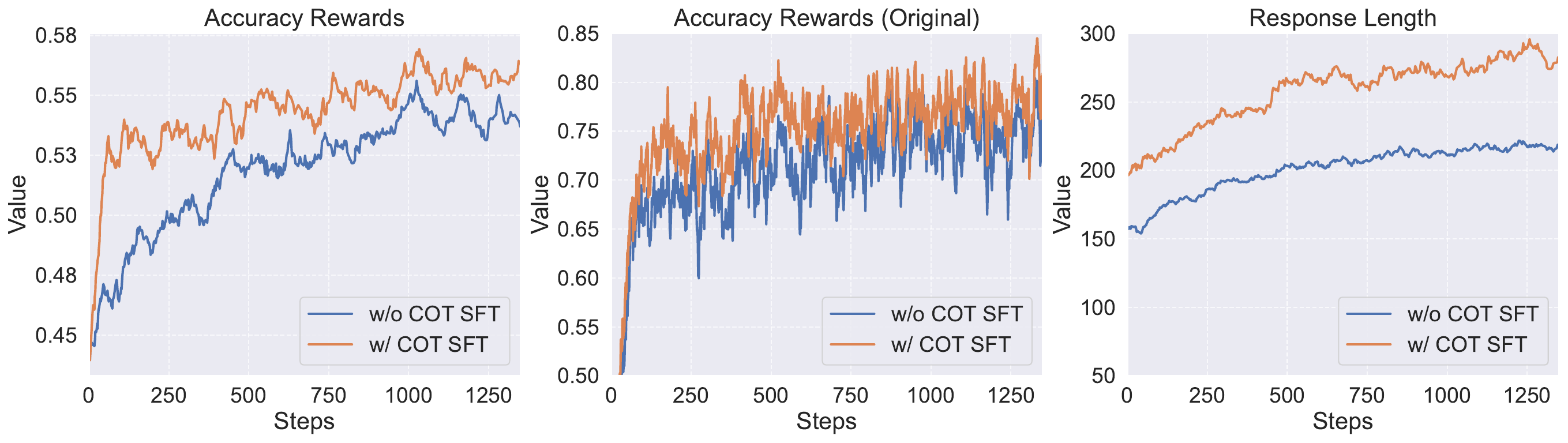}
    \vspace{-5pt}
    \caption{Training curves for the RL stage using the ChartQA and ChartRQA datasets. The orange curve represents our proposed two-stage training strategy, while the blue curve corresponds to a RL-only baseline.}
    \label{fig:rl_vs_cot_rl}
    \vspace{-5pt}
\end{figure*}

\subsection{Main Results}
Table~\ref{sota} shows the performance of Chart-R1 compared with other baseline models. The results show that Chart-R1 achieves the state-of-the-art performance on small-scale ($<$20B) VLMs, including general- and chart-domain models across various chart understanding and reasoning benchmarks.
Especially in ChartQA, Chart-R1 achieves the best performance, even compared with proprietary and large-scale VLMs. In the chart reasoning benchmark, CharXiv-RQ, ChartMuseum and our proposed ChartRQA, Chart-R1 significantly surpass existing chart-domain models. Since the training data of Chart-R1 only contains ChartRQA and ChartQA, these results demonstrate the effectiveness of our proposed ChartRQA dataset and CoT-RL training strategy.

\begin{table*}[t]
  \centering
  \small
  \begin{tabular}{l|cc|ccc}
    \toprule
    \multirow{2}{*}{\textbf{Model Name}} & \multicolumn{2}{c|}{\textbf{Training Setting}}  & \multirow{2}{*}{\textbf{ChartQA}} & \multirow{2}{*}{\textbf{CharXiv-RQ}} & \textbf{ChartRQA} \\
    & SFT & RL & & & (single / multi) \\
    \midrule
    Qwen2.5-VL-7B  &   & & 87.3 & 42.5 & 44.59 / 40.57 \\
    \midrule
    Qwen2.5-VL-7B-SFT  & \textit{QA} &  & 86.16 & 36.0 & 24.76 / 18.34 \\
    \midrule
    \multirow{3}{*}{Qwen2.5-VL-7B-RL}  & & \textit{QA}  & 89.32 & 42.1 & 37.73 / 36.15 \\
    &  &  \textit{QA+RQA-RL} & 90.28 & 45.2 & 44.16 / 40.44\\
    &  \textit{RQA-SFT} & \textit{QA+RQA-RL} & \textbf{91.04} & \textbf{46.2} & \textbf{52.09} / \textbf{49.93} \\
    \bottomrule
  \end{tabular}
  \caption{The ablation study about different SFT and RL training settings. QA and RQA are the abbreviations of ChartQA and ChartRQA.}
  \label{ablation_sft_rl}
\vspace{-5pt}
\end{table*}

\subsection{Ablation Study}
We first assess the impact of different training settings, with results presented in Table~\ref{ablation_sft_rl}. The findings indicate that utilizing our two-stage training strategy
yields the most balanced performance. Notably, omitting Chart-COT causes a significant performance drop on the ChartRQA benchmark. We attribute this to complex charts requiring multi-step thinking before answering. The first Chart-COT stage equips the model with the necessary capability for such step-by-step task decomposition. 
Also, SFT on the ChartQA dataset leads to performance degradation across all benchmarks, including ChartQA itself.
We reckon that although SFT could improve capacity for in-domain tasks, training on simple and low-diversity datasets disrupts the tuned distribution, harming the ability on both in-domain (ChartQA) and OOD (CharXiv-RQ, ChartRQA) tasks.

Prior research underscores the critical role of training data complexity for effective RL \cite{guo2025deepseek}. Our generated ChartRQA training set addresses this by featuring tasks with both single- and multi-chart images, and questions demanding step-by-step reasoning. Including our ChartRQA dataset during the RL stage is crucial for achieving optimal performance. The structural and logical complexity is important for performance enhancements observed in our Chart-RFT stage.
Furthermore, RL exclusively on the ChartQA dataset is insufficient for developing a reasoning model. The limited complexity of ChartQA fails to encourage the model to learn diverse, long-path reasoning strategies. This limitation is empirically demonstrated by the training process shown in Figure~\ref{fig:rl_chartqa}. The accuracy reward rapidly converges to around 0.9 with little subsequent growth, while the response length remains constrained to approximately 100 tokens.

We further investigate the impact of our two-stage training strategy, comparing it to a baseline without the Chart-COT phase. The comparison of RL processes is shown in Figure~\ref{fig:rl_vs_cot_rl}. We find that the first SFT on CoT data has two key benefits. First, it significantly increases the token length generated during the RL phase. Second, it leads to a much effective accuracy reward curve, which rises quickly at the start of training and then converges at a higher final value.

\begin{figure*}[t]
    \centering
    \includegraphics[width=1.0\linewidth]{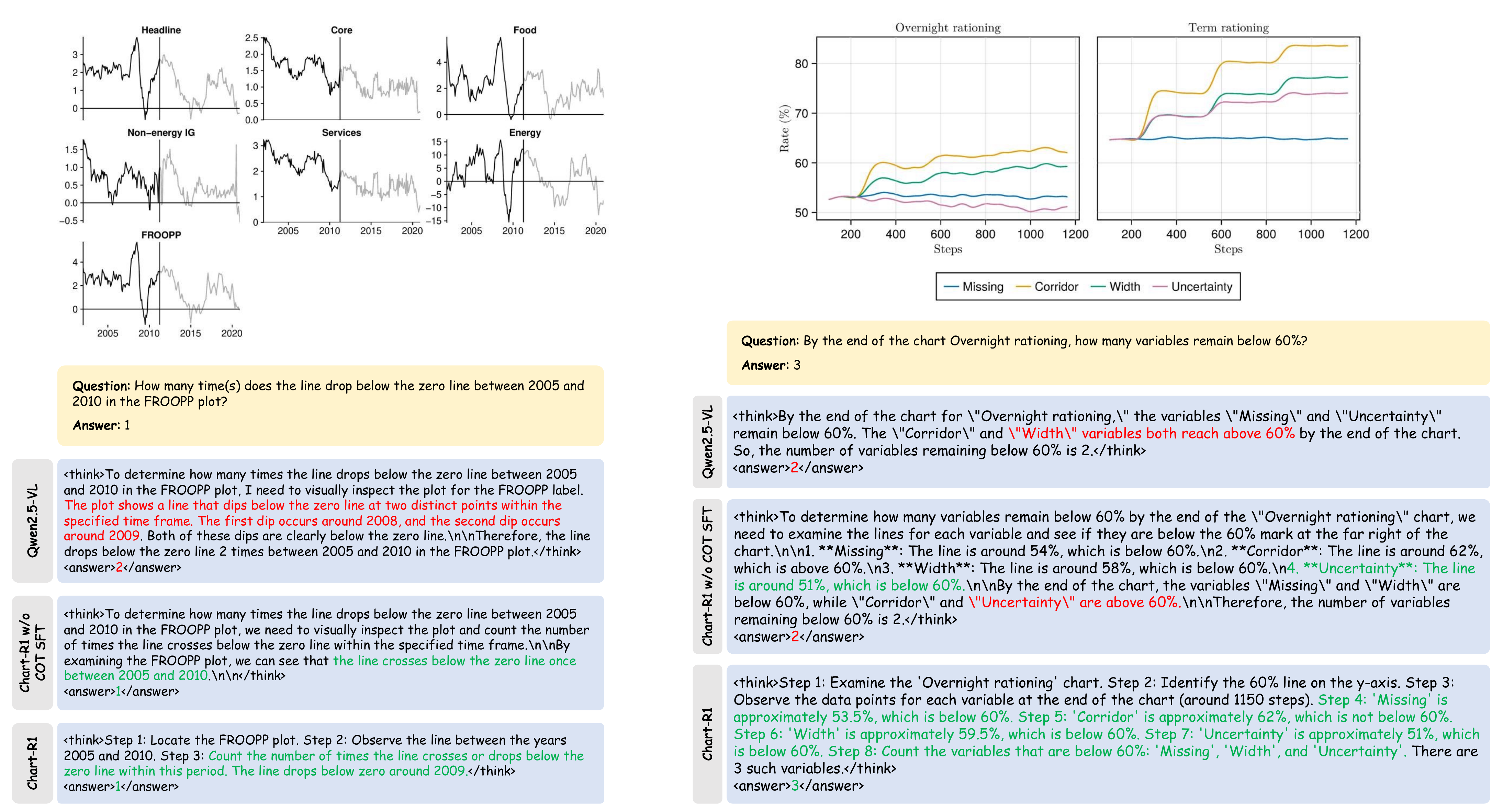}
    \vspace{-15pt}
    \caption{Visualization results of two case studies: (a) Chart-R1 (with and without Chart-COT) both answer correctly while Qwen2.5-VL-7B fails, and (b) only Chart-R1 with Chart-COT answers correctly while both Qwen2.5-VL-7B and Chart-R1 without Chart-COT fail.}
    \label{fig:res_comp_1}
    \vspace{-5pt}
\end{figure*}

\textbf{Visualization} We present qualitative case studies where our Chart-R1 model successfully generates detailed reasoning and correct answers for complex questions in Figure~\ref{fig:res_comp_1}. In these same instances, the baseline Qwen2.5-VL-7B model fails, directly demonstrating the superior performance and more advanced reasoning capabilities of our approach. When Chart-R1 is trained without the Chart-COT stage, it also fails to answer the problems in the right case of Figure~\ref{fig:res_comp_1}. Although it can correctly recognize the chart content, it makes errors during the reasoning process, highlighting the importance of our proposed two-stage training.

\subsection{Error Analysis}
\begin{figure}[t]
    \centering
    \includegraphics[width=1.0\linewidth]{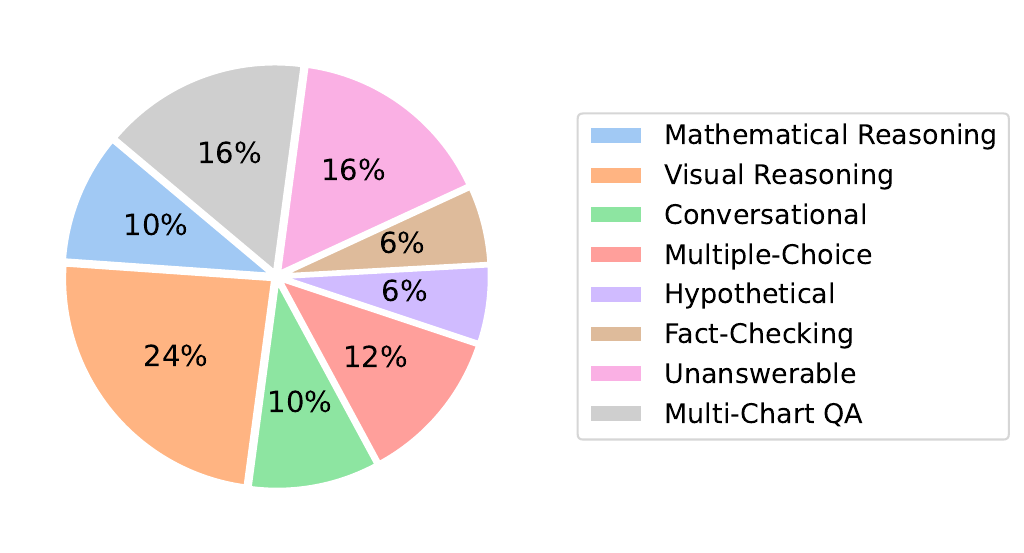}
    \vspace{-30pt}
    \caption{Error type distribution of Chart-R1 on ChartQAPro.}
    \label{fig:error}
    \vspace{-15pt}
\end{figure}

Chart-R1 achieves significant improvements in reasoning ability compared to baseline, but there is still room for further improvement. We randomly sample 50 incorrect responses from Chart-R1 on ChartQAPro and analyze the error types. As shown in Figure~\ref{fig:error}, Chart-R1 is most prone to errors in visual reasoning, multi-chart QA, and unanswerable types. 
Visual reasoning is more challenging than mathematical reasoning, as the latter mainly involves numerical recognition and calculation, while the former requires the model to identify and summarize complex chart patterns.
Multi-chart QA requires the model to integrate information across multiple charts. While ChartRQA was designed to address multi-chart reasoning, the current model still exhibits deficiencies in this aspect. For unanswerable questions, although ChartRQA did not specifically include such samples, Chart-R1 can reject most unanswerable questions through RL, demonstrating good generalizability.

\section{Conclusion}
In this paper, we propose Chart-R1, a chart-domain VLM for complex chart reasoning. To improve the reasoning capacity of Chart-R1, we introduce a programmatic data generation method alongside a novel two-stage training strategy to optimize the data construction and training methodology. Also, we propose ChartRQA, which contains 258k training samples, each constructed in verifiable formats and a benchmark for evaluating complex chart reasoning. The result shows that combining our proposed training strategy, Chart-R1 achieves superior performance compared with other VLMs.

\section*{Limitations}
Our study focuses primarily on statistical charts from academic papers, overlooking practical visualization types such as dashboards and flowcharts. This leads to a gap compared to closed-source models on comprehensive benchmarks such as ChartMuseum. In future research, we plan to expand our training paradigm to incorporate diverse chart types and complex visual reasoning, developing a more versatile chart understanding model.

\section*{Ethical Statement}
Our research employs publicly available models and datasets with proper citations. This approach leverages the widespread use and non-toxic nature of our datasets and prompts to minimize the risk of generating toxic content.

\bibliography{custom}

@article{han2023chartllama,
  title={Chartllama: A multimodal llm for chart understanding and generation},
  author={Han, Yucheng and Zhang, Chi and Chen, Xin and Yang, Xu and Wang, Zhibin and Yu, Gang and Fu, Bin and Zhang, Hanwang},
  journal={arXiv preprint arXiv:2311.16483},
  year={2023}
}

@article{kantharaj2022chart,
  title={Chart-to-text: A large-scale benchmark for chart summarization},
  author={Kantharaj, Shankar and Leong, Rixie Tiffany Ko and Lin, Xiang and Masry, Ahmed and Thakkar, Megh and Hoque, Enamul and Joty, Shafiq},
  journal={arXiv preprint arXiv:2203.06486},
  year={2022}
}

@misc{openai2025o3o4mini,
  author       = {{OpenAI}},
  title        = {Introducing OpenAI o3 and o4-mini},
  year         = {2025},
  month        = {April},
  day          = {16},
  howpublished = {\url{https://openai.com/index/introducing-o3-and-o4-mini/}},
  note         = {Accessed: 2025-07-14}
}

@article{ni2025point,
  title={Point-rft: Improving multimodal reasoning with visually grounded reinforcement finetuning},
  author={Ni, Minheng and Yang, Zhengyuan and Li, Linjie and Lin, Chung-Ching and Lin, Kevin and Zuo, Wangmeng and Wang, Lijuan},
  journal={arXiv preprint arXiv:2505.19702},
  year={2025}
}

@article{masry2024chartgemma,
  title={Chartgemma: Visual instruction-tuning for chart reasoning in the wild},
  author={Masry, Ahmed and Thakkar, Megh and Bajaj, Aayush and Kartha, Aaryaman and Hoque, Enamul and Joty, Shafiq},
  journal={arXiv preprint arXiv:2407.04172},
  year={2024}
}

@article{wang2025multimodal,
  title={Multimodal chain-of-thought reasoning: A comprehensive survey},
  author={Wang, Yaoting and Wu, Shengqiong and Zhang, Yuecheng and Yan, Shuicheng and Liu, Ziwei and Luo, Jiebo and Fei, Hao},
  journal={arXiv preprint arXiv:2503.12605},
  year={2025}
}

@article{wu2025chartcards,
  title={ChartCards: A Chart-Metadata Generation Framework for Multi-Task Chart Understanding},
  author={Wu, Yifan and Yan, Lutao and Shen, Leixian and Mei, Yinan and Wang, Jiannan and Luo, Yuyu},
  journal={arXiv preprint arXiv:2505.15046},
  year={2025}
}

@article{zhao2025chartedit,
  title={ChartEdit: How Far Are MLLMs From Automating Chart Analysis? Evaluating MLLMs' Capability via Chart Editing},
  author={Zhao, Xuanle and Liu, Xuexin and Yang, Haoyue and Luo, Xianzhen and Zeng, Fanhu and Li, Jianling and Shi, Qi and Chen, Chi},
  journal={arXiv preprint arXiv:2505.11935},
  year={2025}
}

@article{tang2025chartmuseum,
  title={ChartMuseum: Testing Visual Reasoning Capabilities of Large Vision-Language Models},
  author={Tang, Liyan and Kim, Grace and Zhao, Xinyu and Lake, Thom and Ding, Wenxuan and Yin, Fangcong and Singhal, Prasann and Wadhwa, Manya and Liu, Zeyu Leo and Sprague, Zayne and others},
  journal={arXiv preprint arXiv:2505.13444},
  year={2025}
}

@article{zhang2024tinychart,
  title={Tinychart: Efficient chart understanding with visual token merging and program-of-thoughts learning},
  author={Zhang, Liang and Hu, Anwen and Xu, Haiyang and Yan, Ming and Xu, Yichen and Jin, Qin and Zhang, Ji and Huang, Fei},
  journal={arXiv preprint arXiv:2404.16635},
  year={2024}
}

@article{xia2024chartx,
  title={Chartx \& chartvlm: A versatile benchmark and foundation model for complicated chart reasoning},
  author={Xia, Renqiu and Zhang, Bo and Ye, Hancheng and Yan, Xiangchao and Liu, Qi and Zhou, Hongbin and Chen, Zijun and Dou, Min and Shi, Botian and Yan, Junchi and others},
  journal={arXiv preprint arXiv:2402.12185},
  year={2024}
}

@article{shi2024chartmimic,
  title={ChartMimic: Evaluating LMM's Cross-Modal Reasoning Capability via Chart-to-Code Generation},
  author={Shi, Chufan and Yang, Cheng and Liu, Yaxin and Shui, Bo and Wang, Junjie and Jing, Mohan and Xu, Linran and Zhu, Xinyu and Li, Siheng and Zhang, Yuxiang and others},
  journal={arXiv preprint arXiv:2406.09961},
  year={2024}
}

@article{wei2022chain,
  title={Chain-of-thought prompting elicits reasoning in large language models},
  author={Wei, Jason and Wang, Xuezhi and Schuurmans, Dale and Bosma, Maarten and Xia, Fei and Chi, Ed and Le, Quoc V and Zhou, Denny and others},
  journal={Advances in neural information processing systems},
  volume={35},
  pages={24824--24837},
  year={2022}
}

@article{shao2024deepseekmath,
  title={Deepseekmath: Pushing the limits of mathematical reasoning in open language models},
  author={Shao, Zhihong and Wang, Peiyi and Zhu, Qihao and Xu, Runxin and Song, Junxiao and Bi, Xiao and Zhang, Haowei and Zhang, Mingchuan and Li, YK and Wu, Y and others},
  journal={arXiv preprint arXiv:2402.03300},
  year={2024}
}

@article{liu2023mmc,
  title={Mmc: Advancing multimodal chart understanding with large-scale instruction tuning},
  author={Liu, Fuxiao and Wang, Xiaoyang and Yao, Wenlin and Chen, Jianshu and Song, Kaiqiang and Cho, Sangwoo and Yacoob, Yaser and Yu, Dong},
  journal={arXiv preprint arXiv:2311.10774},
  year={2023}
}

@article{masry2024chartinstruct,
  title={ChartInstruct: Instruction Tuning for Chart Comprehension and Reasoning},
  author={Masry, Ahmed and Shahmohammadi, Mehrad and Parvez, Md Rizwan and Hoque, Enamul and Joty, Shafiq},
  journal={arXiv preprint arXiv:2403.09028},
  year={2024}
}

@article{masry2022chartqa,
  title={Chartqa: A benchmark for question answering about charts with visual and logical reasoning},
  author={Masry, Ahmed and Long, Do Xuan and Tan, Jia Qing and Joty, Shafiq and Hoque, Enamul},
  journal={arXiv preprint arXiv:2203.10244},
  year={2022}
}

@inproceedings{singh2019towards,
  title={Towards vqa models that can read},
  author={Singh, Amanpreet and Natarajan, Vivek and Shah, Meet and Jiang, Yu and Chen, Xinlei and Batra, Dhruv and Parikh, Devi and Rohrbach, Marcus},
  booktitle={Proceedings of the IEEE/CVF conference on computer vision and pattern recognition},
  pages={8317--8326},
  year={2019}
}

@inproceedings{methani2020plotqa,
  title={Plotqa: Reasoning over scientific plots},
  author={Methani, Nitesh and Ganguly, Pritha and Khapra, Mitesh M and Kumar, Pratyush},
  booktitle={Proceedings of the IEEE/CVF Winter Conference on Applications of Computer Vision},
  pages={1527--1536},
  year={2020}
}

@article{chen2022program,
  title={Program of thoughts prompting: Disentangling computation from reasoning for numerical reasoning tasks},
  author={Chen, Wenhu and Ma, Xueguang and Wang, Xinyi and Cohen, William W},
  journal={arXiv preprint arXiv:2211.12588},
  year={2022}
}

@article{zhao2025chartcoder,
  title={Chartcoder: Advancing multimodal large language model for chart-to-code generation},
  author={Zhao, Xuanle and Luo, Xianzhen and Shi, Qi and Chen, Chi and Wang, Shuo and Liu, Zhiyuan and Sun, Maosong},
  journal={arXiv preprint arXiv:2501.06598},
  year={2025}
}

@article{huang2025vision,
  title={Vision-r1: Incentivizing reasoning capability in multimodal large language models},
  author={Huang, Wenxuan and Jia, Bohan and Zhai, Zijie and Cao, Shaosheng and Ye, Zheyu and Zhao, Fei and Xu, Zhe and Hu, Yao and Lin, Shaohui},
  journal={arXiv preprint arXiv:2503.06749},
  year={2025}
}

@article{meng2025mm,
  title={Mm-eureka: Exploring the frontiers of multimodal reasoning with rule-based reinforcement learning},
  author={Meng, Fanqing and Du, Lingxiao and Liu, Zongkai and Zhou, Zhixiang and Lu, Quanfeng and Fu, Daocheng and Han, Tiancheng and Shi, Botian and Wang, Wenhai and He, Junjun and others},
  journal={arXiv preprint arXiv:2503.07365},
  year={2025}
}

@article{liu2025understanding,
  title={Understanding r1-zero-like training: A critical perspective},
  author={Liu, Zichen and Chen, Changyu and Li, Wenjun and Qi, Penghui and Pang, Tianyu and Du, Chao and Lee, Wee Sun and Lin, Min},
  journal={arXiv preprint arXiv:2503.20783},
  year={2025}
}

@article{yang2025r1,
  title={R1-onevision: Advancing generalized multimodal reasoning through cross-modal formalization},
  author={Yang, Yi and He, Xiaoxuan and Pan, Hongkun and Jiang, Xiyan and Deng, Yan and Yang, Xingtao and Lu, Haoyu and Yin, Dacheng and Rao, Fengyun and Zhu, Minfeng and others},
  journal={arXiv preprint arXiv:2503.10615},
  year={2025}
}

@article{shen2025vlm,
  title={Vlm-r1: A stable and generalizable r1-style large vision-language model},
  author={Shen, Haozhan and Liu, Peng and Li, Jingcheng and Fang, Chunxin and Ma, Yibo and Liao, Jiajia and Shen, Qiaoli and Zhang, Zilun and Zhao, Kangjia and Zhang, Qianqian and others},
  journal={arXiv preprint arXiv:2504.07615},
  year={2025}
}

@article{guo2025deepseek,
  title={Deepseek-r1: Incentivizing reasoning capability in llms via reinforcement learning},
  author={Guo, Daya and Yang, Dejian and Zhang, Haowei and Song, Junxiao and Zhang, Ruoyu and Xu, Runxin and Zhu, Qihao and Ma, Shirong and Wang, Peiyi and Bi, Xiao and others},
  journal={arXiv preprint arXiv:2501.12948},
  year={2025}
}

@article{jia2025chartreasoner,
  title={ChartReasoner: Code-Driven Modality Bridging for Long-Chain Reasoning in Chart Question Answering},
  author={Jia, Caijun and Xu, Nan and Wei, Jingxuan and Wang, Qingli and Wang, Lei and Yu, Bihui and Zhu, Junnan},
  journal={arXiv preprint arXiv:2506.10116},
  year={2025}
}

@article{wang2024charxiv,
  title={Charxiv: Charting gaps in realistic chart understanding in multimodal llms},
  author={Wang, Zirui and Xia, Mengzhou and He, Luxi and Chen, Howard and Liu, Yitao and Zhu, Richard and Liang, Kaiqu and Wu, Xindi and Liu, Haotian and Malladi, Sadhika and others},
  journal={arXiv preprint arXiv:2406.18521},
  year={2024}
}

@article{xu2024chartmoe,
  title={ChartMoE: Mixture of diversely aligned expert connector for chart understanding},
  author={Xu, Zhengzhuo and Qu, Bowen and Qi, Yiyan and Du, Sinan and Xu, Chengjin and Yuan, Chun and Guo, Jian},
  journal={arXiv preprint arXiv:2409.03277},
  year={2024}
}

@article{bai2025qwen25,
  title={Qwen2. 5-vl technical report},
  author={Bai, Shuai and Chen, Keqin and Liu, Xuejing and Wang, Jialin and Ge, Wenbin and Song, Sibo and Dang, Kai and Wang, Peng and Wang, Shijie and Tang, Jun and others},
  journal={arXiv preprint arXiv:2502.13923},
  year={2025}
}

@article{zhu2025internvl3,
  title={Internvl3: Exploring advanced training and test-time recipes for open-source multimodal models},
  author={Zhu, Jinguo and Wang, Weiyun and Chen, Zhe and Liu, Zhaoyang and Ye, Shenglong and Gu, Lixin and Tian, Hao and Duan, Yuchen and Su, Weijie and Shao, Jie and others},
  journal={arXiv preprint arXiv:2504.10479},
  year={2025}
}

@article{masry2025chartqapro,
  title={ChartQAPro: A more diverse and challenging benchmark for chart question answering},
  author={Masry, Ahmed and Islam, Mohammed Saidul and Ahmed, Mahir and Bajaj, Aayush and Kabir, Firoz and Kartha, Aaryaman and Laskar, Md Tahmid Rahman and Rahman, Mizanur and Rahman, Shadikur and Shahmohammadi, Mehrad and others},
  journal={arXiv preprint arXiv:2504.05506},
  year={2025}
}

@inproceedings{hu2024novachart,
  title={Novachart: A large-scale dataset towards chart understanding and generation of multimodal large language models},
  author={Hu, Linmei and Wang, Duokang and Pan, Yiming and Yu, Jifan and Shao, Yingxia and Feng, Chong and Nie, Liqiang},
  booktitle={Proceedings of the 32nd ACM International Conference on Multimedia},
  pages={3917--3925},
  year={2024}
}

@misc{abdin2024phi3technicalreport,
      title={Phi-3 Technical Report: A Highly Capable Language Model Locally on Your Phone}, 
      author={Marah Abdin and Jyoti Aneja and Hany Awadalla and Ahmed Awadallah and Ammar Ahmad Awan and Nguyen Bach and Amit Bahree and Arash Bakhtiari and Jianmin Bao and Harkirat Behl and Alon Benhaim and Misha Bilenko and Johan Bjorck and Sébastien Bubeck and Martin Cai and Qin Cai and Vishrav Chaudhary and Dong Chen and Dongdong Chen and Weizhu Chen and Yen-Chun Chen and Yi-Ling Chen and Hao Cheng and Parul Chopra and Xiyang Dai and Matthew Dixon and Ronen Eldan and Victor Fragoso and Jianfeng Gao and Mei Gao and Min Gao and Amit Garg and Allie Del Giorno and Abhishek Goswami and Suriya Gunasekar and Emman Haider and Junheng Hao and Russell J. Hewett and Wenxiang Hu and Jamie Huynh and Dan Iter and Sam Ade Jacobs and Mojan Javaheripi and Xin Jin and Nikos Karampatziakis and Piero Kauffmann and Mahoud Khademi and Dongwoo Kim and Young Jin Kim and Lev Kurilenko and James R. Lee and Yin Tat Lee and Yuanzhi Li and Yunsheng Li and Chen Liang and Lars Liden and Xihui Lin and Zeqi Lin and Ce Liu and Liyuan Liu and Mengchen Liu and Weishung Liu and Xiaodong Liu and Chong Luo and Piyush Madan and Ali Mahmoudzadeh and David Majercak and Matt Mazzola and Caio César Teodoro Mendes and Arindam Mitra and Hardik Modi and Anh Nguyen and Brandon Norick and Barun Patra and Daniel Perez-Becker and Thomas Portet and Reid Pryzant and Heyang Qin and Marko Radmilac and Liliang Ren and Gustavo de Rosa and Corby Rosset and Sambudha Roy and Olatunji Ruwase and Olli Saarikivi and Amin Saied and Adil Salim and Michael Santacroce and Shital Shah and Ning Shang and Hiteshi Sharma and Yelong Shen and Swadheen Shukla and Xia Song and Masahiro Tanaka and Andrea Tupini and Praneetha Vaddamanu and Chunyu Wang and Guanhua Wang and Lijuan Wang and Shuohang Wang and Xin Wang and Yu Wang and Rachel Ward and Wen Wen and Philipp Witte and Haiping Wu and Xiaoxia Wu and Michael Wyatt and Bin Xiao and Can Xu and Jiahang Xu and Weijian Xu and Jilong Xue and Sonali Yadav and Fan Yang and Jianwei Yang and Yifan Yang and Ziyi Yang and Donghan Yu and Lu Yuan and Chenruidong Zhang and Cyril Zhang and Jianwen Zhang and Li Lyna Zhang and Yi Zhang and Yue Zhang and Yunan Zhang and Xiren Zhou},
      year={2024},
      eprint={2404.14219},
      archivePrefix={arXiv},
      primaryClass={cs.CL},
      url={https://arxiv.org/abs/2404.14219}, 
}

@misc{wu2024deepseekvl2,
      title={DeepSeek-VL2: Mixture-of-Experts Vision-Language Models for Advanced Multimodal Understanding}, 
      author={Zhiyu Wu and Xiaokang Chen and Zizheng Pan and Xingchao Liu and Wen Liu and Damai Dai and Huazuo Gao and Yiyang Ma and Chengyue Wu and Bingxuan Wang and Zhenda Xie and Yu Wu and Kai Hu and Jiawei Wang and Yaofeng Sun and Yukun Li and Yishi Piao and Kang Guan and Aixin Liu and Xin Xie and Yuxiang You and Kai Dong and Xingkai Yu and Haowei Zhang and Liang Zhao and Yisong Wang and Chong Ruan},
      year={2024},
      eprint={2412.10302},
      archivePrefix={arXiv},
      primaryClass={cs.CV},
      url={https://arxiv.org/abs/2412.10302}, 
}

@article{qiu2025metis,
  title={Metis-RISE: RL Incentivizes and SFT Enhances Multimodal Reasoning Model Learning},
  author={Qiu, Haibo and Lan, Xiaohan and Liu, Fanfan and Sun, Xiaohu and Ruan, Delian and Shi, Peng and Ma, Lin},
  journal={arXiv preprint arXiv:2506.13056},
  year={2025}
}

@article{meng2025mmeureka,
      title={MM-Eureka: Exploring the Frontiers of Multimodal Reasoning with Rule-based Reinforcement Learning},
      author={Fanqing Meng and Lingxiao Du and Zongkai Liu and Zhixiang Zhou and Quanfeng Lu and Daocheng Fu and Tiancheng Han and Botian Shi and Wenhai Wang and Junjun He and Kaipeng Zhang and Ping Luo and Yu Qiao and Qiaosheng Zhang and Wenqi Shao},
      year={2025},
      journal={arXiv preprint arXiv:2503.07365},
}

@inproceedings{zheng2024llamafactory,
  title={LlamaFactory: Unified Efficient Fine-Tuning of 100+ Language Models},
  author={Yaowei Zheng and Richong Zhang and Junhao Zhang and Yanhan Ye and Zheyan Luo and Zhangchi Feng and Yongqiang Ma},
  booktitle={Proceedings of the 62nd Annual Meeting of the Association for Computational Linguistics (Volume 3: System Demonstrations)},
  address={Bangkok, Thailand},
  publisher={Association for Computational Linguistics},
  year={2024},
  url={http://arxiv.org/abs/2403.13372}
}

@article{he2024distill,
  title={Distill Visual Chart Reasoning Ability from LLMs to MLLMs},
  author={He, Wei and Xi, Zhiheng and Zhao, Wanxu and Fan, Xiaoran and Ding, Yiwen and Shan, Zifei and Gui, Tao and Zhang, Qi and Huang, Xuanjing},
  journal={arXiv preprint arXiv:2410.18798},
  year={2024}
}

@article{masry2025bigcharts,
  title={BigCharts-R1: Enhanced Chart Reasoning with Visual Reinforcement Finetuning},
  author={Masry, Ahmed and Puri, Abhay and Hashemi, Masoud and Rodriguez, Juan A and Thakkar, Megh and Mahajan, Khyati and Yadav, Vikas and Madhusudhan, Sathwik Tejaswi and Pich{\'e}, Alexandre and Bahdanau, Dzmitry and others},
  journal={arXiv preprint arXiv:2508.09804},
  year={2025}
}

@misc{anthropic2025claude37,
  title = {Claude 3.7 Sonnet and Claude code},
  author = {Anthropic},
  year = {2025},
  month = {February},
  url = {https://www.anthropic.com/news/claude-3-7-sonnet}
}

@misc{kavukcuoglu2025gemini,
  title = {Gemini 2.5: Our most intelligent AI model},
  author = {Kavukcuoglu, Koray},
  year = {2025},
  month = {March},
}

@misc{openai2025gpt41,
  title = {Introducing {GPT-4.1} in the {API}},
  author = {{OpenAI}},
  year = {2025},
  month = {April},
  url = {https://openai.com/index/gpt-4-1/}
}

@misc{bespoke_minichart_7b,
author = {Liyan Tang and Shreyas Pimpalgaonkar and Kartik Sharma and Alexandros G. Dimakis and Maheswaran Sathiamoorthy and Greg Durrett},
title = {Bespoke-MiniChart-7B: pushing the frontiers of open VLMs for chart understanding},     
howpublished ={https://www.bespokelabs.ai/blog/bespoke-minichart-7b}, 
note = {Accessed: 2025-04-23},
year = {2025}
}

@article{zhang2025r1,
  title={R1-vl: Learning to reason with multimodal large language models via step-wise group relative policy optimization},
  author={Zhang, Jingyi and Huang, Jiaxing and Yao, Huanjin and Liu, Shunyu and Zhang, Xikun and Lu, Shijian and Tao, Dacheng},
  journal={arXiv preprint arXiv:2503.12937},
  year={2025}
}

@article{wang2025vl,
  title={Vl-rethinker: Incentivizing self-reflection of vision-language models with reinforcement learning},
  author={Wang, Haozhe and Qu, Chao and Huang, Zuming and Chu, Wei and Lin, Fangzhen and Chen, Wenhu},
  journal={arXiv preprint arXiv:2504.08837},
  year={2025}
}

@article{leng2025mmr1,
  title={Mmr1: Enhancing multimodal reasoning with variance-aware sampling and open resources},
  author={Leng, Sicong and Wang, Jing and Li, Jiaxi and Zhang, Hao and Hu, Zhiqiang and Zhang, Boqiang and Jiang, Yuming and Zhang, Hang and Li, Xin and Bing, Lidong and others},
  journal={arXiv preprint arXiv:2509.21268},
  year={2025}
}

@inproceedings{yang2025effective,
  title={Effective training data synthesis for improving mllm chart understanding},
  author={Yang, Yuwei and Zhang, Zeyu and Hou, Yunzhong and Li, Zhuowan and Liu, Gaowen and Payani, Ali and Ting, Yuan-Sen and Zheng, Liang},
  booktitle={Proceedings of the IEEE/CVF International Conference on Computer Vision},
  pages={2653--2663},
  year={2025}
}

\clearpage

\appendix
\section{Appendix}
\label{sec:appendix}

\subsection{Training Details}
\label{sec:appendix-training}

\noindent \textbf{Chart-COT} We use Qwen2.5-VL-7B-Instruct as the initial model and perform supervised fine-tuning using LLaMA-Factory~\cite{zheng2024llamafactory}. We train the model on the 228k ChartRQA-SFT dataset for one epoch. During training, we freeze the vision tower and multi-modal projector parameters and tune the LLM. The learning rate is set to 1e-5, with a warm-up ratio of 0.1 and batch size of 48. The training process costs 3 hours on 24 H800 GPUs.

\noindent \textbf{Chart-RFT} For the RFT stage, we use the fine-tuned model from the Chart-COT stage. We adopt the MM-EUREKA~\cite{meng2025mmeureka} framework based on OpenRLHF for training. The model is trained for 3 episodes using 30k ChartQA and 30k ChartRQA-RL. We set the rollout batch size and the training batch size to 128, with each sample generating 8 rollouts. The temperature for model generation is set to 1, and we exclude KL divergence in the loss calculation. The learning rate is set to 1e-6, with a warm-up ratio of 0.03, while freezing the vision tower during training. Following the default setting for instruction models, the format reward coefficient is set to 0.5. We employ the online filtering strategy with lower and upper bounds of 0.1 and 0.9, respectively. The training process costs 30 hours on 24 H800 GPUs.

\subsection{Benchmark Details}
\textbf{ChartQA}~\cite{masry2022chartqa} focuses on chart question answering with complex reasoning questions that involve logical and arithmetic operations. Following the settings in the original paper, we evaluate models on the test set reporting overall accuracy scores across both human-written (ChartQA-H) and machine-generated (ChartQA-M) question subsets.

\noindent \textbf{CharXiv}~\cite{wang2024charxiv} presents a comprehensive evaluation suite with natural, challenging, and diverse charts from arXiv papers to provide a more realistic assessment of chart understanding capabilities. We evaluate models on the Reasoning Questions (CharXiv-RQ) subset, which requires synthesizing information across complex visual elements in charts. Following the original paper, we use GPT-assisted evaluation to assess model responses.

\noindent \textbf{ChartQAPro}~\cite{masry2025chartqapro} introduces a diverse benchmark with various chart types, including infographics and dashboards, and question formats that better reflect real-world challenges. We evaluate models using Chain-of-Thought (CoT) prompting in the original paper and report overall accuracy across five question types.

\noindent \textbf{ChartMuseum}~\cite{tang2025chartmuseum} is a chart question-answering benchmark designed to evaluate complex visual reasoning capabilities with expert-annotated questions from diverse real-world charts. Following the original paper, we evaluate models using the provided CoT prompt and LLM-as-a-Judge evaluation method.

\subsection{More Ablation Studies}

\begin{table*}[t]
  \centering
  \small
  \begin{subtable}[t]{0.45\linewidth}
    \centering
    \begin{tabular}{l|cc}
      \toprule
      \textbf{RL Setting} & \textbf{ChartQA} & \textbf{CharXiv-RQ} \\
      \midrule
      \multicolumn{3}{c}{\textit{Accuracy Reward}} \\
      \midrule
      ED & 89.88 & 44.0 \\
      ED + SM & \textbf{90.28} & \textbf{45.2} \\
      \midrule
      \multicolumn{3}{c}{\textit{RL Training Set}} \\
      \midrule
      QA & 89.32 & 42.1 \\
      QA + RQA$^{\dag}$ & \textbf{90.32} & 44.6 \\
      QA + RQA & 90.28 & \textbf{45.2} \\
      \bottomrule
    \end{tabular}
    \caption{Ablation study on reward and training set in RL.}
    \label{ablation_reward}
  \end{subtable}
  \hfill
  \begin{subtable}[t]{0.5\linewidth}
    \centering
    \begin{tabular}{l|cc}
      \toprule
      \textbf{SFT Setting} & \textbf{ChartQA} & \textbf{CharXiv-RQ} \\ 
      \midrule
      \multicolumn{3}{c}{\textit{SFT Training Set}} \\
      \midrule
      RQA-SFT\&RL + QA & 88.40 & 41.2 \\
      RQA-SFT & \textbf{89.88} & \textbf{44.5} \\
      \midrule
      \multicolumn{3}{c}{\textit{SFT Dataset}} \\
      \midrule
      TinyChart & 84.80 & 36.1 \\
      ChartGemma & 86.72 & 39.1 \\
      ChartRQA-SFT & \textbf{90.20} & \textbf{45.0} \\
      \bottomrule
    \end{tabular}
    \caption{Ablation study on training set and different datasets in SFT.}
    \label{ablation_sft_data}
  \end{subtable}
  \caption{Ablation studies within RL and SFT stages. ED and SM are the abbreviations of Edit Distance and Soft Matching. RQA$^{\dag}$ indicates that only samples with chart types of line, bar, and pie in ChartRQA are used for training.}
  \label{tab:ablation_combined}
  \vspace{-10pt}
\end{table*}

\textbf{Reward Function} To assess different accuracy rewards, we conduct experiments by training Qwen2.5-VL-7B-Instruct for the RL stage only, as shown in Table~\ref{ablation_reward}. The results demonstrate that employing a soft accuracy reward, which combines edit distance for string-based tasks and soft matching for numerical tasks, yields superior performance across both benchmarks. This finding underscores the importance of adjusting the reward function to the specific type of answers.

\noindent \textbf{Image Diversity \& Question Complexity} Our ChartRQA dataset is characterized by two key features: diverse chart images and complex questions requiring step-by-step reasoning. To investigate the importance of these factors in RL training, we select samples from ChartRQA-RL that only include line, bar, and pie chart types, which are the same chart types found in the ChartQA dataset, and train the model using RL only. As shown in Table~\ref{ablation_reward}, without increasing chart type diversity, the complex questions in ChartRQA still substantially enhance the model's reasoning ability. Furthermore, using the full ChartRQA dataset, which includes a wider variety of chart images, leads to further improvements on CharXiv-RQ.

\noindent \textbf{SFT Data Composition} When training Chart-R1, our SFT dataset consists of 228k samples from our ChartRQA-SFT. We then ablate the SFT data composition by adding two sources, the ChartQA dataset and the 30k ChartRQA-RL that overlaps with the RL data, to assess the impact on performance. We train each setting for 2k steps and 1 epoch for SFT and RL, respectively. The results in Table~\ref{ablation_sft_data} show that combining ChartQA and ChartRQA-RL, the final performance decreases evidently. Our analysis indicates that using overlapping data for SFT and RL leads to overfitting, where the model memorizes reasoning paths from the SFT stage, resulting in more rigid thinking processes and a significant loss of output diversity.
Also, the direct-answer format of ChartQA discourages the model from developing the ability to break down problems into step-by-step thinking process.

\noindent \textbf{Comparison with Existing Chart Datasets} To enable a fair comparison between ChartRQA and existing chart SFT datasets, we replace the SFT dataset with TinyChart and ChartGemma, while keeping all other settings consistent. TinyChart is a comprehensive dataset that integrates multiple open-source datasets and comprises a variety of tasks. To ensure that the model focuses on chart understanding and reasoning, we exclude the Chart-to-text and Chart-to-table generation tasks. For the RL stage, we use a combination of ChartQA and ChartRQA-RL for 1 epoch of training. 
As shown in Table~\ref{ablation_sft_data}, Chart-R1 trained on ChartRQA-SFT achieves the best performance on both benchmarks. The results indicate that the unified thinking and answer format and the effective step-by-step reasoning process in ChartRQA are key factors in enhancing the model's reasoning ability.

\subsection{ChartRQA Analysis}
\label{sec:data_detail}

\begin{figure*}[t]
    \centering
    \includegraphics[width=0.95\linewidth]{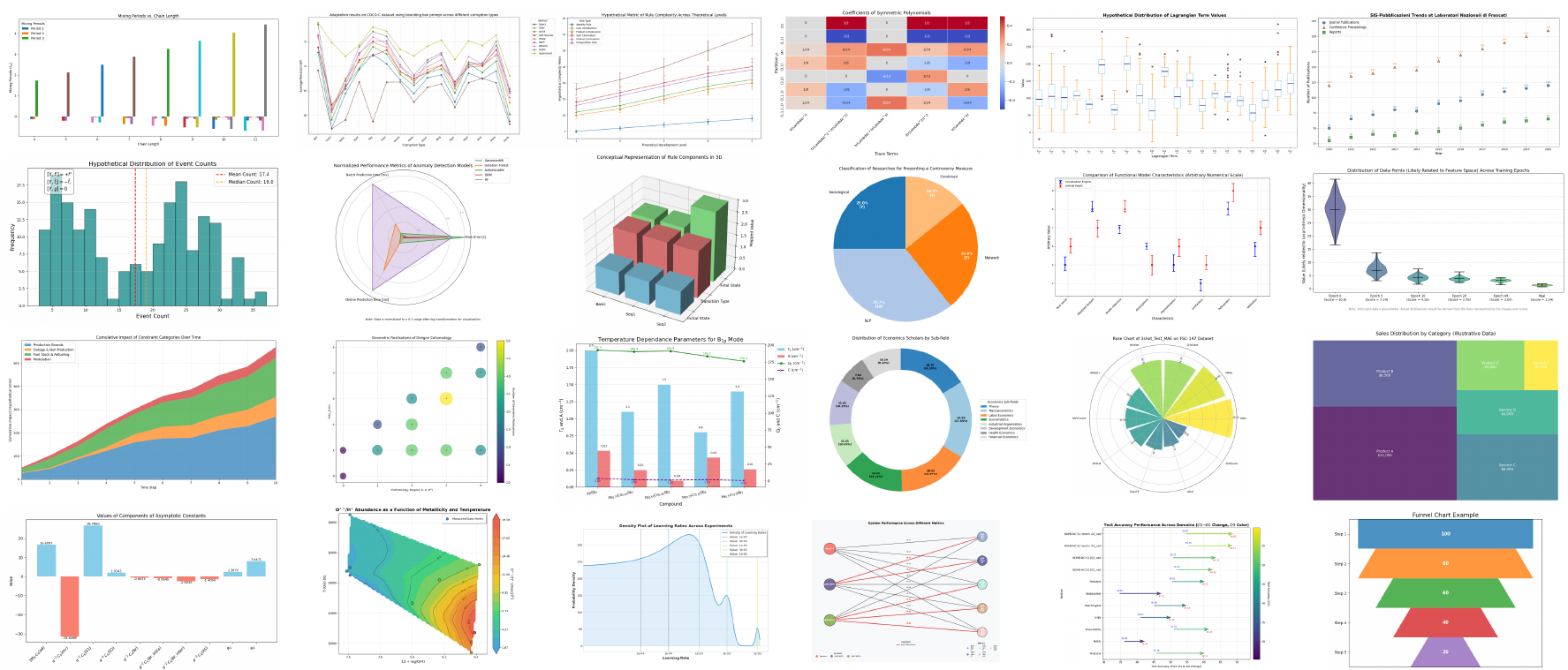}
    \caption{Single-chart samples of 24 chart types from ChartRQA.}
    \label{fig:single_chart_type}
\end{figure*}

\begin{figure*}[!htbp]
    \centering
    \includegraphics[width=0.95\linewidth]{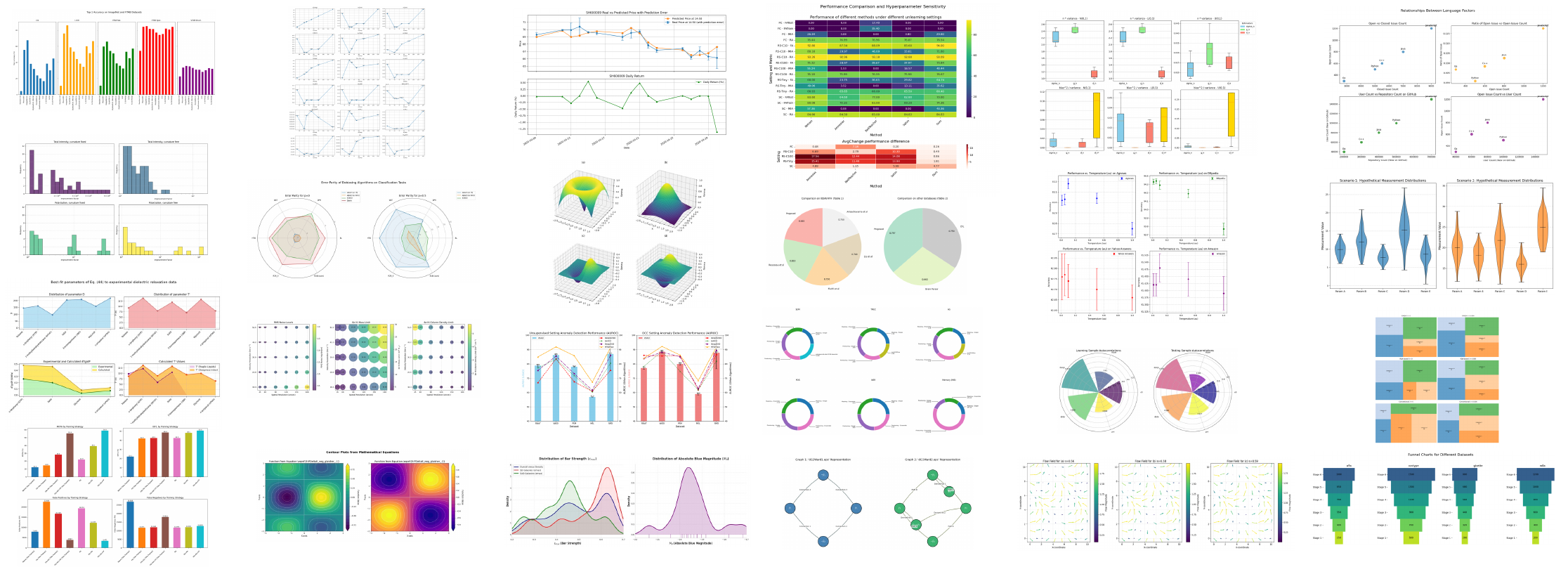}
    \caption{Multi-chart samples of 24 chart types from ChartRQA.}
    \label{fig:multi_chart_type}
\end{figure*}

We count the quantity and distribution of different chart types across the training and test sets of ChartRQA, as detailed in Table~\ref{chart_types}. The distribution among the various types to be well-balanced.
Furthermore, Figures~\ref{fig:single_chart_type} and ~\ref{fig:multi_chart_type} provide visualization examples of 24 chart types from the ChartRQA dataset, showcasing both single-chart and multi-chart formats, respectively.

\begin{table*}[t]
  \centering
  \small
  \begin{tabular}{c|ccccccccc}
    \toprule
    Split & Bar & Line & ErrorBar & Heatmap & Box & Scatter & Histogram & Radar & 3D \\
    \midrule
    Train & 11,850 & 10,752 & 11,838 & 8,993 & 12,112 & 10,299 & 15,856 & 9,483 & 9,746 \\
    Test & 100 & 88 & 83 & 60 & 103 & 76 & 116 & 46 & 65 \\
    \midrule
    Split & Pie & ErrorPoint & Violin & Area & Bubble & Multi-axes & Ring & Rose & Treemap \\
    \midrule    
    Train & 17,812 & 10,814 & 12,571 & 9,175 & 8,996 & 10,776 & 12,726 & 10,533 & 9,850 \\
    Test & 103 & 68 & 116 & 75 & 51 & 61 & 54 & 61 & 64 \\
    \midrule
    Split & Bar\_num & Contour & Density & Graph & Quiver & Funnel & Total \\
    \midrule
    Train & 12,150 & 10,291 & 12,860 & 8,764 & 9,955 & 227 & 258,429 \\
    Test  & 64 & 67 & 77 & 47 & 52 & 5 & 1,702 \\
    \bottomrule
  \end{tabular}
  \caption{The detailed chart types and corresponding quantities in our proposed ChartRQA train and test set. ChartRQA contains 24 chart types, each of which contains approximate samples.}
  \label{chart_types}
\end{table*}

\subsection{Prompts}
\label{sec:prompt}
\begin{figure*}[t]
    \centering
    \includegraphics[width=0.95\linewidth]{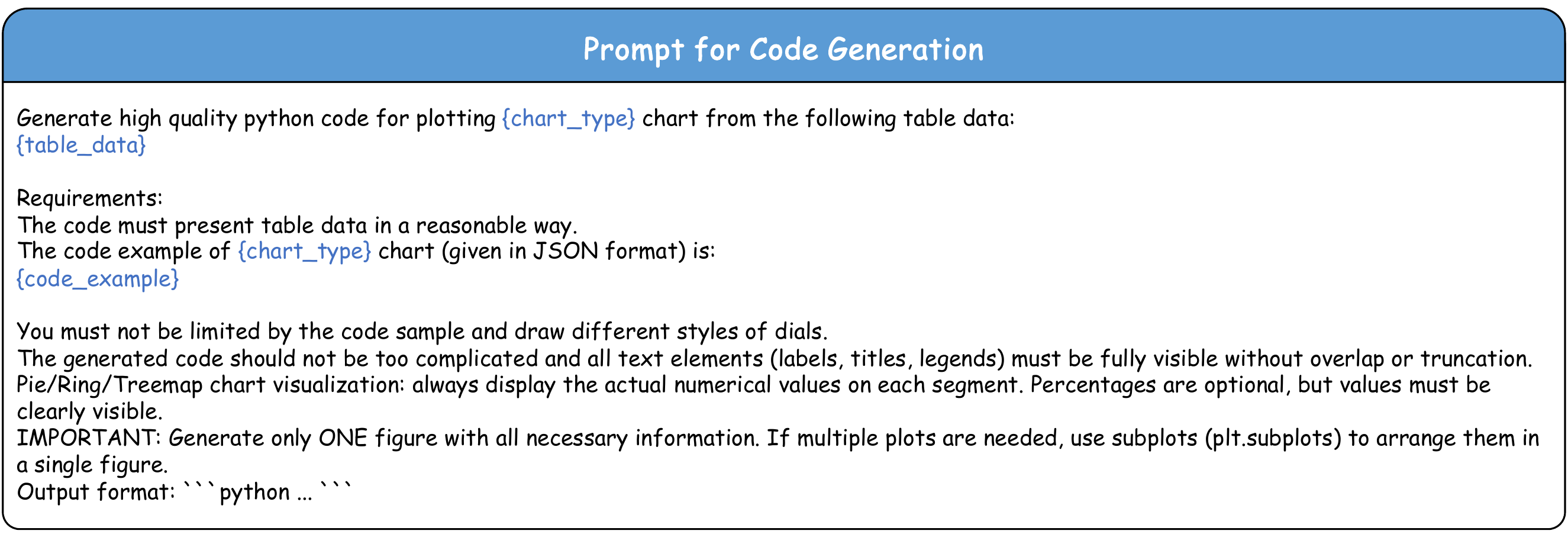}
    \caption{Prompt for code generation.}
    \label{fig:prompt_code_gen}
\end{figure*}

\begin{figure*}[t]
    \centering
    \includegraphics[width=0.95\linewidth]{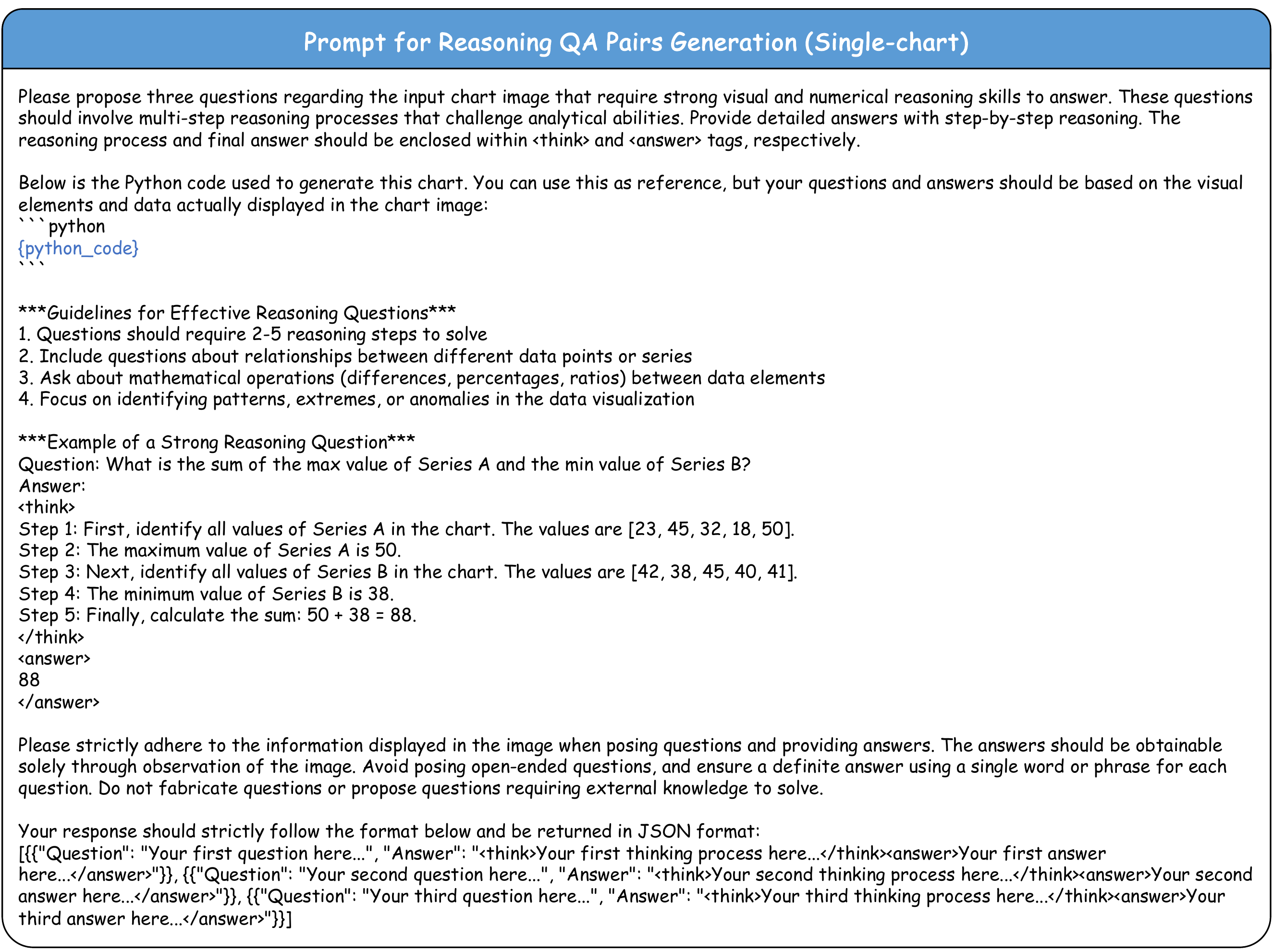}
    \caption{Prompt for reasoning QA pairs generation for single-chart formats.}
    \label{fig:prompt_single_qa}
\end{figure*}

\begin{figure*}[t]
    \centering
    \includegraphics[width=0.95\linewidth]{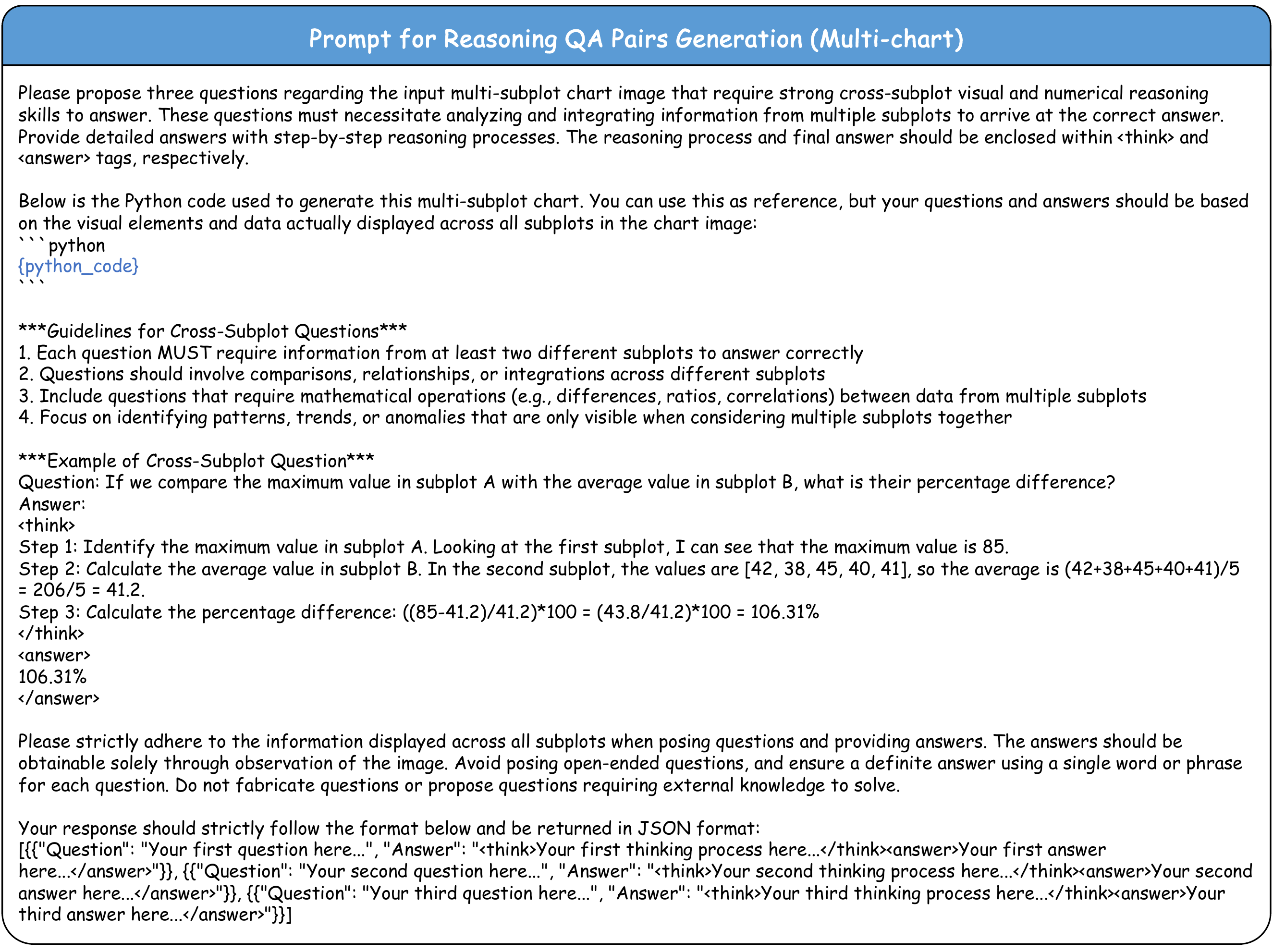}
    \caption{Prompt for reasoning QA pairs generation for multi-chart formats.}
    \label{fig:prompt_multi_qa}
\end{figure*}

\begin{figure*}[t]
    \centering
    \includegraphics[width=0.95\linewidth]{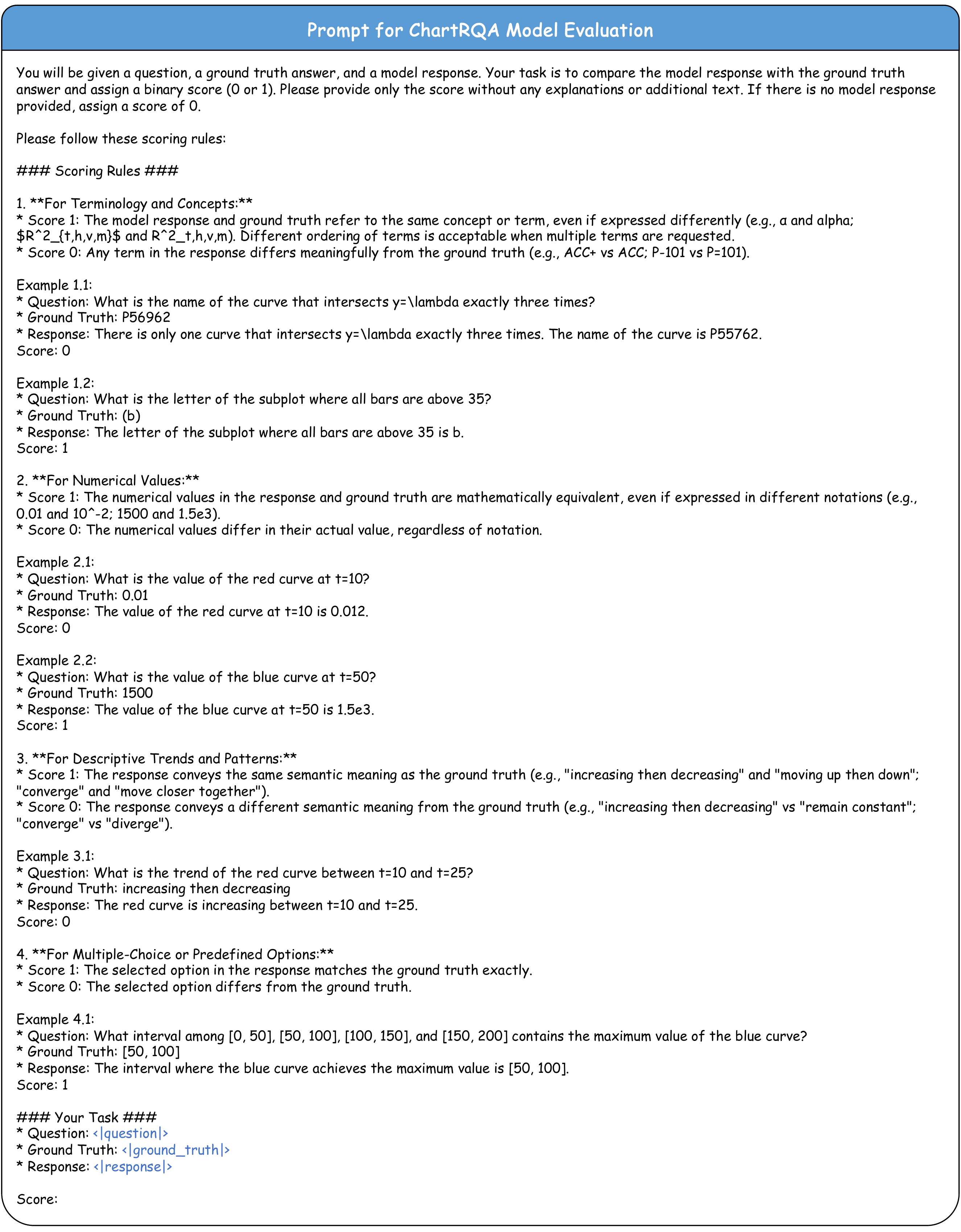}
    \caption{Prompt for ChartRQA evaluation using GPT-4o.}
    \label{fig:prompt_model_eval}
\end{figure*}

To enhance transparency and reproducibility, we provide the exact prompts used for dataset generation and evaluation.
For data generation, we employ Gemini-2.5-Flash to generate both plotting code and QA pairs for data construction. Figure~\ref{fig:prompt_code_gen} illustrates the prompt used for plotting code generation. We utilize real table data as input, select one chart type from the 24 predefined chart types, and sample a code example corresponding to that chart type to generate the plotting code. 
Figures~\ref{fig:prompt_single_qa} and ~\ref{fig:prompt_multi_qa} display the prompts used to generate reasoning QA pairs for single-chart and multi-chart formats, respectively. We craft an example for each format to aid LLMs in understanding complex chart reasoning tasks and to generate step-by-step reasoning processes and precise answers that conform to the format. The executable plotting code is provided as auxiliary information to LLMs, making the generated QA pairs more reliable.
Figure~\ref{fig:prompt_model_eval} shows the prompt used for model evaluation. We employ GPT-4o to assess the match between the ground truth and the model's predictions, where GPT-4o returns a score of 0 or 1 to indicate the correctness of the model's prediction. Our evaluation focuses solely on the correctness of the final answer, disregarding the reasoning process.

\subsection{The Use of AI Assistants}
In this work, we used LLMs as writing tools to improve language clarity and readability. These models helped refine the text and enhance the presentation of our ideas. All research concepts, experiments, implementations, and analyses were conducted independently by the authors.

\end{document}